# A Practical Cross-Device Federated Learning Framework over 5G Networks

Wenti Yang, Naiyu Wang, Zhitao Guan*, Longfei Wu, Xiaojiang Du, Mohsen Guizani

*Abstract*—The concept of federated learning (FL) was first proposed by Google in 2016. Thereafter, FL has been widely studied for the feasibility of application in various fields due to its potential to make full use of data without compromising the privacy. However, limited by the capacity of wireless data transmission, the employment of federated learning on mobile devices has been making slow progress in practical. The development and commercialization of the 5th generation (5G) mobile networks has shed some light on this. In this paper, we analyze the challenges of existing federated learning schemes for mobile devices and propose a novel cross-device federated learning framework, which utilizes the anonymous communication technology and ring signature to protect the privacy of participants while reducing the computation overhead of mobile devices participating in FL. In addition, our scheme implements a contribution-based incentive mechanism to encourage mobile users to participate in FL. We also give a case study of autonomous driving. Finally, we present the performance evaluation of the proposed scheme and discuss some open issues in federated learning.

## Ⅰ. INTRODUCTION

Nowadays, mobile devices are widely used in various application scenarios, as a result, a huge amount of data is generated. The artificial intelligence (AI) technology can make the data more valuable. Specifically, the data can be used to train machine learning (ML) models and these models can be used to improve the applications and services related to mobile devices. For example, optimizing wireless networks (e.g., content caching, spectrum management, 5G core network) [1], advancing intelligence of Internet of Vehicles, and so on. As we all know, one of the most important factors for training a good machine learning model is to employ a large amount of real data. The data collected from a single mobile device is limited and biased. Therefore, it should be enabled for multiple devices to share their data for the training of machine learning models together. The sharing of large amounts of raw data may leads to two main challenges: privacy leakage and excessive communication overhead. Although existing technologies such as anonymous communication, differential privacy, and public key encryption, can be used to alleviate the risk of privacy leakage to some extent, the later cannot be solved at the same time [2].

Federated learning (FL) is a promising AI technology to solve the above problems. However, in practical, due to the limited capacity of wireless communication and computing power of mobile devices, the application of federated learning for mobile devices is restricted [3]. The emergence of 5G networks brings new opportunities for FL on mobile devices. The FL coupled with the fast and reliable 5G wireless communications is ideal for the secure and practical data sharing among mobile devices [4].

### 1.1. Introduction to Federated Learning

Federated Learning is a distributed ML technology that provides privacy-preservation. In FL, multiple participants collaborate to train a ML model, with the participants' raw data kept locally to themselves. In the server-client based horizontal FL, each participant uses local data to train the model and uploads the model parameters to an aggregation server. The server is responsible for aggregating the model parameters uploaded by each participant, generating the global model parameters and returning them to each participant. The above process iterates until the model parameters converge or meet the preset conditions.

In terms of training samples, the types of federated learning mainly include horizontal FL and vertical FL. The Horizontal FL is for horizontally partitioned data which has the same feature space but different sample spaces. For example, the same type of information from different users in different banks. The vertical FL is for vertically partitioned data which has the same sample space but different feature spaces. For example, different types of information from the same user in a bank and in a medical care system. In this paper, we mainly talk about the horizontal FL.

In terms of application scenarios, the types of federated learning mainly include cross-device FL and cross-silo FL. The cross-device FL is usually used in mobile device applications and has the characteristics of a large number of participants with a small amount of raw data owned by each participant. In contrast to the cross-device FL, only several reliable organizations are involved in cross-silo FL. In this paper, we mainly study the cross-device FL. [5]

Besides, in addition to the server-client FL, some researchers proposed peer-to-peer (P2P) FL. The key idea of P2P FL is to avoid the potentially untrusted third party by using P2P communication between the peer participants. However, the excessive communication overhead has become a huge obstacle to P2P FL.

In addition to using the cross-device FL to optimize 5G wireless communications, there are many potential application scenarios of cross-device FL in the context of 5G networks, such as autonomous driving, vehicle to everything, medical care, smart grid and other IoT-based applications [4].

### 1.2. Challenges of Cross-Device FL and Our Motivation

Although the development of 5G networks makes it possible for federated learning to be efficiently carried out between mobile devices, there are still some challenges:

1) Privacy leakage

Researchers found that the output vectors, model parameters, gradients of ML model may reveal sensitive information of the training data and the parameters of the model. In the application process of ML models, there are some attacks (e.g.,

* Zhitao Guan is the corresponding author (email: guan@ncepu.edu.cn)
Wenti Yang, Zhitao Guan and Naiyu Wang are with School of Control and Computer Engineering, North China Electric Power University
Longfei Wu is with Department of Mathematics & Computer Science, Fayetteville State University
Xiaojiang Du is with Department of Electrical and Computer Engineering, Stevens Institute of Technology, Hoboken NJ, USA
Mohsen Guizani is with Department of Computer Science and Engineering, Qatar University

model extraction attack, model inversion attack, membership-inference attack) that may cause the leakage of model parameters or the training data [5]. In the training process of FL, participants need to send the updated gradients or the model parameters of each iteration to the server or other participants, which may also reveal the private information of the training data. Privacy leakage is still a challenge that cannot be ignored in FL.

2) Unreliable mobile devices with limited computing power

Most existing FL schemes use the following privacy preservation techniques to solve the above privacy leakage problem during the FL training process:

   a) Pairwise additive masking: Adding masks to the local gradients and model parameters is a commonly used privacy preservation technique in FL. To prevent data from becoming unavailable due to the superimposition of masked data during the aggregation process, participants need to interact multiple times to eliminate the mask in the aggregation [6]. This restricts the participants from withdrawing halfway. However, mobile devices participating in FL may fail or drop out due to various reasons such as network disruption, low battery, and so on. Therefore, this method may lead to poor robustness of the system.

   b) Differential privacy: Using differential privacy to add noise to the local gradients and model parameters is also a good solution to protect privacy. However, mobile devices usually have less training data, and adding noise may cause data to be inefficient [7].

   c) Secure multi-party computation: Some researches proposed the FL privacy preservation schemes based on secure multi-party computation such as garbled circuits, homomorphic encryption and secret sharing, which aggregate the gradients and parameters in the form of encrypted circuits or ciphertexts. [8]. This method is computationally expensive, and not suitable for devices with limited computing power (e.g., mobile devices).

In general, the above methods are suitable for the cross-silo FL with only a few stable participants: each participant has sufficient computing capability, large training data set and the communication between participants is stable, for example, FL for several banks. On the contrary, most mobile devices rely on wireless networks for communication and are often widely distributed. In addition, the computing capability of mobile devices is limited and the training data set is relatively small. Therefore, a more practical privacy preservation scheme for mobile devices is needed.

3) Incentive and fairness

Due to the concerns about privacy leakage or simply the unwillingness to devote computing resources, mobile users may be reluctant to participate in the federated learning. In addition, participants with different contributions to the model are rewarded with the same global model parameters, which may discourage the active participants. In order to motivate more mobile users to participate and ensure fairness, a reasonable incentive mechanism needs to be added. The existing incentive mechanisms for FL mainly include: contribution-based incentive mechanism [9], reputation-based incentive mechanism and resource allocation incentive mechanism [10,11]. There are also some FL schemes that consider both privacy preservation and performance when implementing the incentive mechanism [10]. These schemes use game theory, blockchain and other technologies to achieve novel incentive mechanisms. However, most of them motivate users by monetary reward, ignoring the role of the models. In addition, some incentive mechanisms mainly focus on resource allocation, how to quantify the value of participants' local data (e.g., data quality and data quantity) for FL training privately, and ensure the fairness of FL is also a challenge. [12]

To address the above challenges, we propose a practical cross-device federated learning framework and give a case study on autonomous driving. Our framework has the following features:

   a) It adopts the anonymous communication technology, participants do not need to interact multiple times and waste additional computing resources, which can provide privacy preservation while reducing the computational overhead.

   b) Considering that there may be adversaries posing as participants to affect the training of the model, we adopt the ring signature to verify the identities of participants.

   c) It utilizes a contribution-based incentive mechanism that can quantify the value of participants' training data privately, different from existing incentive mechanisms that uses monetary rewards, model-based rewards can help to improve and/or optimize the mobile applications and services on their devices.

## II. THE GENERAL FRAMEWORK

### 2.1. Description of the Proposed Framework

Our FL framework contains two layers: the local training layer and the aggregation layer, as shown in Fig. 1. In the local training layer, participants (mobile devices) use their local data to update the global model and get different local models. In the aggregation layer, the aggregation server aggregates the local models uploaded by participants to generate/update the global model. This is an iterative process, the detailed steps are as follows:

   a) The aggregation server sets a unified initial global model and distributes the model parameters to the mobile devices participating in FL.

   b) Each participant contributes to the global model by training its own local data and generates the local model. Since participants use their own local data for the training, their trained local models vary from each other.

   c) Each participant then uploads the local model parameters to the aggregation server through the anonymous communication network. As a result, the aggregation server and the adversary cannot find out the true identities of the owners of the local model parameters collected in a certain iteration.

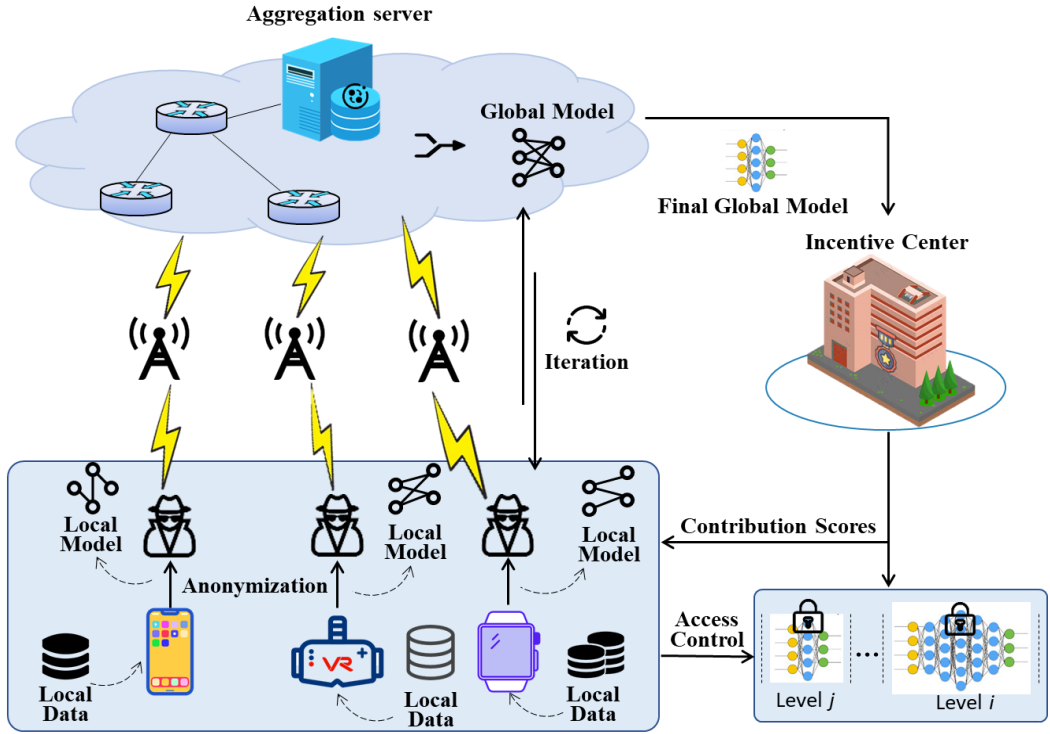

**Figure** 1. The proposed framework

d) The aggregation server aggregates the local model parameters uploaded by the participants and generates/updates the global model. There are multiple aggregation rules, such as federated averaging (FedAvg), centroid distance weighted FedAvg, and so on. The aggregation server distributes the global model parameters to the participants. When the global model parameters converge or meet the preset requirements, the iteration terminates. Otherwise, repeat steps b-d.

e) The aggregation server sends the final global model to the incentive center.

f) The incentive center distributes the contribution scores for each participant and ranks each global model according to the preset's rules.

g) Participants with a certain degree of contribution can access the corresponding model. This step is implemented by an access control scheme.

**2.2. Characteristics of the Proposed Framework**

Considering the features of the mobile device environment, our proposed federated learning framework for mobile devices has the following characteristics.

1) Privacy-preservation

As we mentioned earlier that privacy leakage could happen not only in the application of the model, but also in the training and iteration of FL. The main focus of this paper is to prevent the privacy leakage in the training and iteration of FL. To protect the privacy of participants, the anonymous communication technology and ring signature is adopted. When a participant submits updated model parameters to the aggregation server, the network address can be anonymized to protect the participants' identities. In this way, the adversaries cannot figure out who the model parameters belong to, so that they cannot infer participants' privacy through the process of iterative updates.

2) Trade-off between privacy and computation overhead

In the mobile device environment, due to the limited computing capability, reducing the computation overhead is always a top priority. Therefore, we abandon the conventional privacy preservation methods, such as homomorphic encryption (HE), and instead make a trade-off between the level of privacy and the computation overhead.

In terms of the computation overhead, we take the HE-based FL scheme as an example. Before uploading the model parameters to the aggregation server, participants need to encrypt each element in the vectors of the parameters using HE, so that the parameters can be aggregated in the form of ciphertext. In some deep neural network models, the number of elements can reach the size of millions, which means each participant needs to perform millions of encryption operations. Although some works have proposed the batch encryption, there is still a limitation on the number of parameters that can be encrypted at one time [8]. In this paper, the ring signature we use does not need to calculate the elements one by one, instead, it uses cascade or hash function to map all elements to one element to achieve verification.

In terms of privacy, under the security guarantee of encryption techniques, the server or the adversaries cannot obtain any information about the parameters. In our paper, due to the use of anonymous communication technology, neither the server nor the adversaries can know the owner of the parameters in each round. Therefore, even if the server receives the parameters, it cannot infer any private information as the belongings of these values are unknown. Besides, the ring signature used in our scheme can prevent the adversaries from posing as participants and submitting fake parameters.

3) System robustness

Since the large number of participants in the cross-device federated learning can be widely distributed, the communication among the participants may be unstable. The privacy-preservation methods such as adding mask require multiple rounds of interactions among the peer participants to prevent data from becoming unavailable. However, participants' intentional or unintentional withdrawals may affect the accuracy of the final model. In our FL framework, participants only need to communicate with the server, which has better tolerance for single-device failure or disconnection.

4) Incentive

Due to the concerns about privacy leakage, or simply the unwillingness to devote computing resources, mobile users may be reluctant to participate in the federated learning. In order to get more mobile devices to participate the FL, we propose a contribution- based incentive mechanism with access control. Different from existing incentive mechanisms, sharing the outcome of federated learning – the global model with the participants can help to improve and/or optimize the mobile applications and services on their devices. This can motivate mobile users to actively participate and provide high quality data to the FL. Avoiding monetary incentives can also prevent some legal issues.

## III. CASE STUDY: APPLICATION ON AUTONOMOUS DRIVING

In recent years, autonomous driving has made some progress. However, it is still a huge challenge for autonomous driving to deal with complex and unforeseen environments. One of the main reasons is that the amount of training samples used for autonomous driving learning algorithms is not sufficient. Federated learning, as a promising solution, can use the actual data collected from each autonomous car for model training while protecting the privacy of each individual participant.

### 3.1. Security Assumption

First, we define the security of the system according to the actual conditions in autonomous driving.

1) Participant: We assume that the participants (autonomous cars) are honest-but-curious. They do not submit fake model parameters maliciously, but they may try to figure out the private information of other participants.

2) Aggregation server: The aggregation server is a semi-honest third party, it may return wrong aggregation results due to laziness and is curious about the privacy of the participants.

3) Malicious adversary: The malicious adversaries may try to recover the participants' private information from their model parameters. Additionally, the malicious adversaries may impersonate legitimate participants to send fake model parameters to the aggregation server to corrupt the global model.

4) Key generation center (KGC): The key generation center is responsible for generating the system parameters of ring signature. It is a fully trusted party and does not participate in the training of FL. After it generates the system parameters, it goes offline.

### 3.2. Preliminary

1) Ring Signature

The ring signature is a digital signature scheme that can achieve the anonymity of the signer's identity. The core idea of the ring signature is that there are n users, and each user has a public key and a private key. When a user signs a message m, it needs to use the public key of other users and his own private key to generate the signature. The verifier can verify that the signature is generated by one of the n signers, but the actual signer cannot be located. A user can choose any possible set of signers to produce a valid ring signature, and use the public key of these signers and his own private key to complete the signing operation. We adopt ring signature to prevent adversaries from masquerading as the legitimate participants.

2) Homomorphic Hash

The homomorphic hash function is a kind of collision-resistant hash function satisfying the homomorphic property. Given an additive homomorphic hash function $H$, several random numbers $a_1, a_2, ..., a_n$. According to the data field of the corresponding hash function, the value of $H(a_1) + H(a_2) + ... + H(a_n)$ is equal to the value of $H(a_1 + a_2 + ... + a_n)$. This special hash function can verify the correctness of the calculation result without knowing the raw data. The verifier only needs to obtain the hash value of each parameter to verify whether the calculation result (the sum of these parameters) is correct. Our framework conducts the correctness verification for the aggregation results using homomorphic hash.

### 3.3. Description of the scheme

1) System Initialization

When a new FL task needs to be initiated, the KGC first uses the number of participants $n$ and the aggregation server to decide a unified initial model and parameters. The KGC needs to generate the following parameters.

- A pair of public and private keys ($pk_i$, $sk_i$) for each participant $i$, which is used for the ring signature.
- A hash function $H$ with homomorphic property which is used to verify the correctness of the aggregation results.

The KGC sends the key pair ($pk_i$, $sk_i$) to the corresponding participant $i$ and announces $pk_i$ and the hash function $H$ to all participants and the aggregation server.

2) Local Training

The local training phase includes the following steps.

- Each participant downloads the unified initial model and parameters (represented by $\omega_0$) from the aggregation server.

- Each participant $i$ uses the local data sets and parameters $\omega_0$ to perform the local model training operation: $LocalUpdate(i, \omega_0) \rightarrow \omega_{1,i}$. Due to the different local data sets used for training, the local model constructed by each participant is different.

- Each participant $i$ calculates the ring signature of the local model parameters, as shown in Fig. 2: $RSign(\omega_{1,i}, pk_i, pk_j, ..., pk_k, sk_i) \rightarrow R_{\omega(1,i)}$. The participant $i$ can select the public keys for the ring signature from all the $n$ participants (The greater the number of public keys selected, the better for the privacy-preservation, but the greater the computation overhead), and use the public keys as well as his own private key $sk_i$ to sign the parameters $\omega_{1,i}$. Then it uploads the local parameters $\omega_{1,i}$, the ring signature $R_{\omega(1,i)}$ and the signers' public keys used for ring signature to the aggregation server via the anonymous communication network. In order to verify the correctness of the aggregation results, each participant $i$ calculates the homomorphic hash of the local model parameters $H(\omega_{1,i})$, and multicasts $H(\omega_{1,i})$ to other participants.

- If the local training is not the first round, after downloading the parameters $\omega_k$ from the aggregation server, the participants first verify whether the aggregation results calculated by the aggregation server are correct. Each participant only needs to verify whether the sum of the homomorphic hash values $H(\omega_{k,i})$ of all participants is equal to $H(\omega_k)$.

3) Aggregation

In the aggregation phase, the aggregation server needs to execute two tasks.

First, the aggregation server verifies if the identities of the participants sending the parameters are legitimate.

We assume that *VSign* is the verification algorithm for the ring signature. The aggregation server can use $\omega_{1,i}$, $R_{\omega(1,i)}$ and the public keys to verify whether $VSign(\omega_{1,i}, pk_i, pk_j, ..., pk_k, R_{\omega(1,i)})$ is equal to $\omega_{1,i}$. This verification can prevent malicious adversaries other than the n participants from submitting fake model parameters. During this process, the aggregation server cannot figure out the identity of the uploader of the model parameters, so the identities of the participants are protected.

Then, the aggregation server aggregates the local parameters uploaded by each participant.

The aggregation server averages the local model parameters $\omega_{1,i}$ of all the n participants and produces the updated global model parameters $\omega_1$. Then the aggregation server checks whether the updated global model parameters $\omega_1$ have converged. If they have converged, the federated training ends and the final global model parameters are obtained. Otherwise, each participant downloads $\omega_1$ from the aggregation server and the training process will repeat.

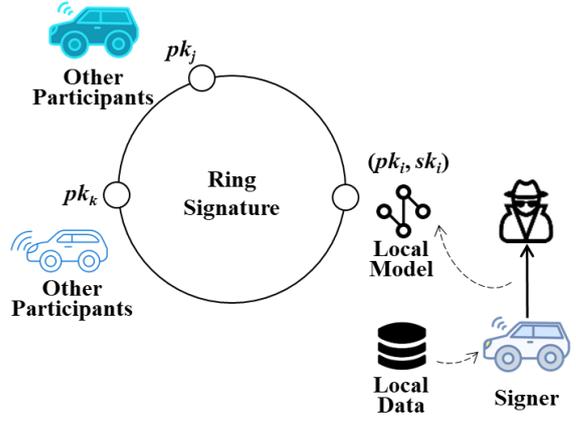

**Figure 2.** Ring signature

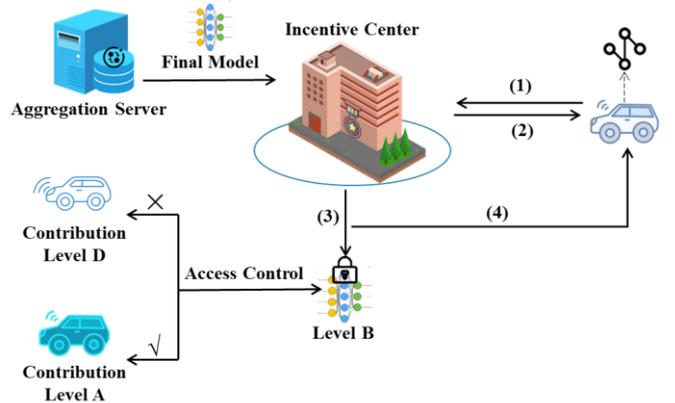

**Figure 3.** Contribution-based incentive mechanism

4) Incentive

The incentive mechanism is shown in Fig.3. Its detailed description is as follows.

Before participating in the federated learning training, the user first uses his local data to train a local machine learning model and proves the quality of the model to the incentive center in the manner of zero-knowledge. The user holds the model, and the incentive center holds the data to be inferred. They perform secure inference over the ML model using secure 2-party computation (2PC) such as oblivious transfer (OT) and garbled circuits [13]. 2PC can ensure that the incentive center cannot obtain the user's model parameters, so as to protect the privacy of the user. The user also cannot get the inferred data, and thus the deliberately modified inference results become meaningless (step 1). The incentive center distributes a contribution weight ε to the user according to the inference accuracy, which, to a certain extent, characterizes the possible contribution that the user's data made to the corresponding FL model (step 2).

After the federated learning task is completed, a global model will be generated. The incentive center seeks the users' consent, and adds tags to the trained model, grades it according to the usage, accuracy, and so on. Assuming that these models are classified into four levels: A, B, C and D. The incentive center encrypts the model to implement the access control scheme. Each level corresponds to an attribute. For example, the access policy of the D-level model is set to A or B or C or D, and the access policy of the B-level model is set to A or B. Only users who have reached the corresponding contribution level can decrypt the model. The encrypted models are stored in the model market (step 3). The participants in the model can gain credits of contribution according to the number of times the model is accessed, the model level and their respective contribution weight ε (step 4).

After reaching a certain contribution level, the user can request for a secret key SK of the access control scheme from KGC to access models of the same level. In the access control scheme, even if the users are of the same level, the granted secret keys are different, which can prevent the abuse of the secret key.

## IV. PERFORMANCE EVALUATION

In this part, we evaluate the performance of our proposed framework from two perspectives: (1) the impact of the number of participants on accuracy and (2) when there is an adversary, the impact of verification of participants on accuracy.

We build the federated learning environment with Python (version 3.6.2) and TensorFlow (version 2.3.2). A multi-layer perceptron model is conducted as the experimental subject for the training on MNIST dataset of handwritten digits with a training set of 60,000 examples, and a test set of 10,000 examples. From Fig. 4, we can conclude that when a user conducts learning only based on her own local data, the accuracy of the model is much lower than that of the FL. Additionally, when the sample number of each user is certain, the more users involved in learning, the higher accuracy the overall global model can achieve - this also highlights the importance of using the incentive mechanism.

In the case study, We use the ring signature to protect privacy and verify the identity of participants to prevent adversaries from impersonating legitimate participants. Here we construct a malicious adversary in the experiment to show the importance of verifying participants. As shown in Fig.5, in the FL scheme without the verification function, a malicious adversary may greatly downgrade the accuracy of the model.

To better show the advantages of our scheme in terms of computation overhead, we also give a comparison of the computation overhead between the privacy preservation technique used in our scheme (i.e., ring signature) and the HE commonly used in FL.

We choose paillier as the HE algorithm, the encryption time of each parameter is about 0.037s. For the linear regression model, assuming that the feature dimension is 10, the computation overhead of encryption for each participant in one round is 0.407s. For a fully connected layer with 300 input neurons and 100 output neurons, the number of parameters can reach 330,100 and the corresponding computation overhead is 12213.7s. Similarly, some practical convolution layers also have tens of thousands of parameters. At present, the parameters of some popular deep neural networks can reach

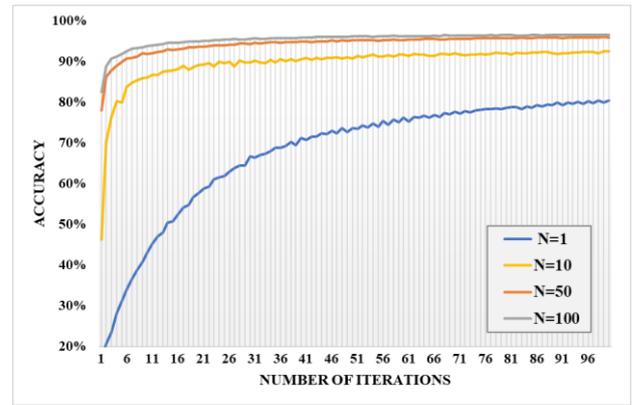

Figure 4. The effect of the number of participants on accuracy. N is the number of participants in FL.

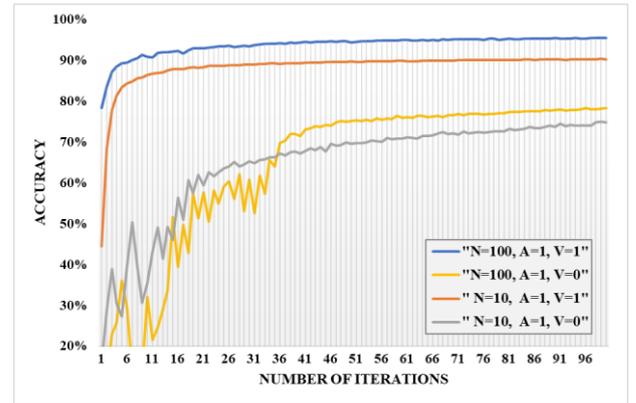

Figure 5. The effect of verification of participants on accuracy. A is the number of adversaries, V=1 means that there is a verification process, otherwise, V=0.

the level of one million or even hundreds of millions, in which the HE can hardly work.

The computation overhead of ring signature used in our scheme is only related to the number of public keys used for signatures, for example, 2 public keys for 0.0165s, 10 public keys for 0.0192s and 100 public keys for 0.056s.

## V. OPEN ISSUES

1) How to solve the problem that participants may submit fake parameters

To the best of our knowledge, none of the existing studies has successfully solved the problem of participants submitting fake parameters. Some related studies [14] have tried to solve this issue. However, these studies only judge whether the participant is honest based on the parameters submitted by the participant. It may make a misjudgment, causing injustice to the honest participants and may lead to overfitting of the model (The model performed well on the raw data set, but poorly on the new data set). Zero-knowledge proof may be a promising solution.

2) How to realize efficient federated learning for vertically partitioned data

Most of the current federated learning schemes are for horizontally partitioned data, but there are few studies on federated learning for vertically partitioned data. It is relatively difficult to implement federated learning for vertically partitioned data [15]. However, in the cross-device FL, there are some scenarios that require vertically federated learning, for example, medical data and traffic data of the same user. It is necessary to carry out more in-depth research on vertically federated learning.

## VI. CONCLUSION

In this paper, we employed the anonymous communication technology to construct a cross-device federated learning framework based on 5G mobile networks. Our framework has lower computation overhead while protecting the privacy of mobile users. We give a case study of autonomous driving. The ring signature is used to verify the identity of participants and the hash homomorphism is used for the correctness verification for the calculation results of the aggregation server. In addition, we implemented a contribution-based incentive mechanism with access control to encourage mobile users to participate in federated learning. The performance evaluation proves the practicality of our scheme. Finally, we discussed some open issues in federated learning.

## ACKNOWLEDGMENT

The work is partially supported by the National Natural Science Foundation of China under Grant 61972148.

## BIOGRAPHIES

WENTI YANG (yangwt@ncepu.edu.cn) is currently a PhD candidate at the School of Control and Computer Engineering, North China Electric Power University (NCEPU). She received her B.S. and M.S. degree from NCEPU in 2017 and 2021. Her current research focuses on privacy preserving machine learning.

NAIYU WANG [Graduate Student Member] (naiyuwang@ncepu.edu.cn) is currently pursuing the master's degree with the School of Control and Computer Engineering, North China Electric Power University. Her current research interests include blockchain and applied cryptography.

ZHITAO GUAN [M'13] (guan@ncepu.edu.cn) is currently an Associate Professor at the School of Control and Computer Engineering, North China Electric Power University. He received his BEng degree and PhD in Computer Application from Beijing Institute of Technology, China, in 2002 and 2008, respectively. His current research focuses on smart grid security, wireless security and cloud security. He has authored over 60 peer-reviewed journal and conference papers in these areas. He is a Member of the IEEE.

LONGFEI WU (lwu@uncfsu.edu) is currently an assistant professor in the Department of Mathematics and Computer Science at Fayetteville State University. He received his Ph.D. degree in computer and information sciences from Temple University in July 2017. He obtained his B.E. degree from Beijing University of Posts and Telecommunications in July 2012. His research interests are the security and privacy of networked systems and modern computing devices, including mobile devices, IoT, implantable medical devices, and wireless networks.

XIAOJIANG (JAMES) DU [S'99–M'03–SM'09–F'20] (dxj@ieee.org) is the Anson Wood Burchard Endowed-Chair Professor in the Department of Electrical and Computer Engineering at Stevens Institute of Technology. He was a professor at Temple University. He received his B.S. from Tsinghua University, Beijing, China in 1996. He received his M.S. and Ph.D. degrees in electrical engineering from the University of Maryland, College Park in 2002 and 2003, respectively. His research interests are security, wireless networks, and systems. He has authored over 500 journal and conference papers in these areas, as well as a book published by Springer. He has been awarded more than 8 million U.S. dollars in research grants from the U.S. National Science Foundation (NSF), Army Research Office, Air Force Research Lab, the State of Pennsylvania, and Amazon. He won the best paper award at IEEE ICC 2020, IEEE Globecom 2014, and the best poster runner-up award at ACM MobiHoc 2014. He serves on the editorial boards of three IEEE journals. He is an IEEE Fellow and a Life Member of ACM.

MOHSEN GUIZANI [S'85-M'89–SM'99–F'09] (mguizani@ieee.org) received the B.S. and M.S. degrees in electrical engineering, the M.S. and Ph.D. degrees in computer engineering from Syracuse University, Syracuse, NY, USA, in 1984, 1986, 1987, and 1990, respectively. He is currently a Professor at the Computer Science and Engineering Department in Qatar University, Qatar. His research interests include wireless communications and mobile computing, computer networks, mobile cloud computing, security, and smart grid. He is currently the Editor-in-Chief of the IEEE Network Magazine, serves on the editorial boards of several international technical journals and the Founder and Editor-in-Chief of Wireless Communications and Mobile Computing journal. He is the author of nine books and more than 500 publications in refereed journals and conferences. He is a Fellow of IEEE and a Senior Member of ACM.